\documentclass[11pt,a4paper]{article}
\usepackage[hyperref]{acl2018}
\usepackage{times}
\usepackage{latexsym}
\usepackage{url}  %Required
\usepackage{graphicx}  %Required
\usepackage{booktabs}
\usepackage{amssymb}
\usepackage{amsmath}
\usepackage{multirow}
\usepackage{verbatim}
\usepackage[utf8x]{inputenc}
\aclfinalcopy % Uncomment this line for the final submission
\title{Neural Models for Key Phrase Extraction and Question Generation}

\author{Sandeep Subramanian$^{\spadesuit\clubsuit}$ ~ Tong Wang$^{\clubsuit}$ ~ Xingdi Yuan$^{\clubsuit}$ ~ Saizheng Zhang$^{\spadesuit}$ \\ ~ \textbf{Yoshua Bengio $^{\spadesuit\dagger}$ ~ Adam Trischler{$^\clubsuit$}}\\
   $^\clubsuit$Microsoft Research, Montréal ~~ $^\spadesuit$MILA, Université de Montréal ~~ $^\dagger$CIFAR Senior Fellow \\
   { \tt sandeep.subramanian.1@umontreal.ca}
}

\date{}

\begin{document}
\maketitle
\begin{abstract}
We propose a two-stage neural model to tackle question generation from documents. First, our model estimates the probability that word sequences in a document are ones that a human would pick when selecting candidate answers by training a neural key-phrase extractor on the answers in a question-answering corpus. Predicted key phrases then act as target answers and condition a sequence-to-sequence question-generation model with a copy mechanism. Empirically, our key-phrase extraction model significantly outperforms an entity-tagging baseline and existing rule-based approaches. We further demonstrate that our question generation system formulates fluent, answerable questions from key phrases. This two-stage system could be used to augment or generate reading comprehension datasets, which may be leveraged to improve machine reading systems or in educational settings.
\end{abstract}

\section{Introduction}
Question answering and machine comprehension has gained increased interest in the past few years. An important contributing factor is the emergence of several large-scale QA datasets~\cite{rajpurkar2016squad,trischler2016,nguyen2016,joshi2017triviaqa}. However, the creation of these datasets is a labour-intensive and expensive process that usually comes at significant financial cost. Meanwhile, given the complexity of the problem space, even the largest QA dataset can still exhibit strong biases in many aspects including question and answer types, domain coverage, linguistic style, etc.

To address this limitation, we propose and evaluate neural models for automatic question-answer pair generation that involves two inter-related components: first, a system to identify candidate answer entities or events (key phrases) within a passage or document \cite{becker2012mind}; second, a question generation module to construct questions about a given key phrases. As a financially more efficient and scalable alternative to the human curation of QA datasets, the resulting system can potentially accelerate further progress in the field.

Specifically, We formulate the key phrase extraction component as modeling the probability of potential answers conditioned on a given document,~i.e., $P(a|d)$. Inspired by successful work in question answering, we propose a sequence-to-sequence model that generates a set of key-phrase \emph{boundaries}. This model can flexibly select an arbitrary number of key phrases from a document. To teach it to assign high probability to human-selected answers, we train the model on large-scale, crowd-sourced question-answering datasets.

We thus take a purely data-driven approach to understand the priors that humans have when selecting answer candidates, working from the premise that crowdworkers tend to select entities or events that interest them when formulating their own comprehension questions. If this premise is correct, then the growing collection of crowd-sourced question-answering datasets~\cite{rajpurkar2016squad,trischler2016} can be harnessed to learn models for key phrases of interest to human readers.

Given a set of extracted key phrases, we then approach the question generation component by modeling the conditional probability of a question given a document-answer pair,~i.e., $P(q|a,d)$. To this end, we use a sequence-to-sequence model with attention \cite{DBLP:journals/corr/BahdanauCB14} and the pointer-softmax mechanism \cite{gulcehre2016pointing}. This component is also trained to maximize the likelihood of questions estimated on a QA dataset. When training this component, the model sees the ground truth answers from the dataset.

Empirically, our proposed model for key phrase extraction outperforms two baseline systems by a significant margin. We support these quantitative findings with qualitative examples of generated question-answer pairs given documents.

\section{Related Work}

\subsection{Key Phrase Extraction}
An important aspect of question generation is identifying which elements of a given document are important or interesting to inquire about.
Existing studies formulate key-phrase extraction in two steps. In the first, lexical features (e.g., part-of-speech tags) are used to extract a key-phrase candidate list exhibiting certain types \cite{liu2011gap,wang2016ptr,le2016unsupervised,yang2017semisupervisedqa}. In the second, ranking models are often used to select a phrase from among the candidates. \citet{medelyan2009human_competitive,lopez2010humb} used bagged decision trees, while \citet{lopez2010humb} used a Multi-Layer Perceptron (MLP) and
Support Vector Machine to perform binary classification on the candidates. \citet{Mihalcea2004textrank,wan2008neighborhood_knowledge,le2016unsupervised} scored key phrases using \textit{PageRank}. \citet{heilman2010rating} asked crowdworkers to rate the acceptability of computer-generated \emph{natural language} questions as quiz questions, and \citet{becker2012mind} solicited quality ratings of text chunks as potential gaps for \emph{Cloze-style} questions.

These studies are closely related to our proposed work by the common goal of modeling the distribution of key phrases given a document. The major difference is that previous studies begin with a prescribed list of candidates, which might significantly bias the distribution estimate. In contrast, we adopt a dataset that was originally designed for question answering, where crowdworkers presumably tend to pick entities or events that interest them most. We postulate that the resulting distribution, learned directly from data, is more likely to reflect the true relevance of potential answer phrases.

Recently, \citet{meng2017deepkpg} proposed a generative model for key phrase prediction with an encoder-decoder framework that is able both to generate words from a vocabulary and point to words from the document. Their model achieved state-of-the-art results on multiple keyword-extraction datasets. This model shares certain similarities with our key phrase extractor, i.e., using a single neural model to learn the probabilities that words are key phrases. Since their focus was on a hybrid abstractive-extractive task in contrast to the purely extractive task in this work, a direct comparison between works is difficult.

\citet{yang2017semisupervisedqa} used rule-based methods to extract potential answers from unlabeled text, and then generated questions given documents and extracted answers using a pre-trained question generation model. The model-generated questions were then combined with human-generated questions for training question answering models. Experiments showed that question answering models can benefit from the augmented data provided by their approach.

\subsection{Question Generation}
Automatic question generation systems are often used to alleviate (or eliminate) the burden of human generation of questions to assess reading comprehension \cite{mitkov2003computer,kunichika2004automated}. Various NLP techniques have been adopted in these systems to improve generation quality, including parsing \cite{heilman2010good,mitkov2003computer}, semantic role labeling \cite{lindberg2013generating}, and the use of lexicographic resources like \emph{WordNet} \cite{miller1995wordnet,mitkov2003computer}. However, the majority of the proposed methods resort to simple, rule-based techniques such as template-based slot filling \cite{lindberg2013generating,chali2016ranking,labutov2015deep} or syntactic transformation heuristics \cite{agarwal2011automatic,ali2010automation} (e.g., subject-auxiliary inversion, \cite{heilman2010good}). These techniques generally do not capture the diversity of human generated questions.

To address this limitation, end-to-end-trainable neural models have recently been proposed for question generation in both vision \cite{mostafazadeh2016generating} and language. For the latter, \citet{du2017learning} used a sequence-to-sequence model with an attention mechanism derived from the encoder states. \citet{yuan2017machine} proposed a similar architecture but further improved model performance with policy gradient techniques. \citet{wang2017jointqa} proposed a generative model that learns jointly to generate questions or answers from documents.

\section{Model Description}
\subsection{Notations}
Several components introduced in the following sections share the same model architecture for encoding text sequences. The common notations are explained in this section.

Unless otherwise specified, $w$ refers to word tokens, $\boldsymbol e$ to word embeddings and $\boldsymbol h$ to the annotation vectors (also commonly referred to as hidden states) produced by an RNN. Superscripts specify the source of a word, e.g., $d$ for documents, $p$ for key phrases, $a$ for (gold) answers, and $q$ for questions. Subscripts index the position inside a sequence. For example, $\boldsymbol e^d_i$ is the embedding vector for the $i$-th token in the document.

A sequence of words are often encoded into \textit{annotation vectors} (denoted $\boldsymbol h$) by applying a bidirectional LSTM encoder to the corresponding sequence of word embeddings. For example, $\boldsymbol h^q_j=\mathrm{LSTM}(\boldsymbol e^q_j,\boldsymbol h^q_{j-1})$ is the annotation vector for the $j$-th word in a question.

\subsection{Key Phrase Extraction}
In this section, we describe a simple baseline as well as two proposed neural models for extracting key phrases (answers) from documents.

\subsubsection{Entity Tagging Baseline}
As our first baseline, we use \textit{spaCy}\footnote{https://spacy.io/docs/usage/entity-recognition} to predict all entities in a document as relevant key phrases (call this model \textit{ENT}). This is motivated by the fact that entities constitute the largest proportion (over 50\%) of answers in the SQuAD dataset \cite{rajpurkar2016squad}. Entities includes dates (\textit{September 1967}), numeric entities (\textit{3, five}), people (\textit{William Smith}), locations (\textit{the British Isles}), and other named concepts (\textit{Buddhism}).

\subsubsection{Neural Entity Selection} The baseline model above na{\"i}vely selects all entities as candidate answers. One pitfall is that it exhibits high recall at the expense of precision (Table~\ref{tab:da_table}), since not all entities lead to interesting questions. We first attempt to address this with a neural entity selection model (\texttt{NES}) that selects a subset of entities from a list of candidates provided by our \texttt{ENT} baseline. Our neural model takes as input a document (i.e., a sequence of words), $D = (w^d_1,\dots,w^d_{n_d})$, and a list of $n_e$ entities as a sequence of (\textit{start}, \textit{end}) locations within the document, $E = ((e^{start}_1, e^{end}_1), \dots, (e^{start}_{n_e}, e^{end}_{n_e}))$. The model is then trained on the binary classification task of predicting whether an entity overlaps with any of the human-provided answers.

Specifically, we maximize $\sum_{i}^{n_e} \log(P(e_i|D))$. We parameterize $P(e_i|D)$ using a three-layer multilayer perceptron (MLP) that takes as input the concatenation of three vectors $\langle \boldsymbol h_{n_d}^d;\boldsymbol h_{avg}^d;\boldsymbol h_{e_i}\rangle$. Here, $\boldsymbol h_{avg}^d$ and $\boldsymbol h_{n_d}^d$ are the average and the final state of the document annotation vectors, respectively, and $\boldsymbol h_{e_i}$ is the average of the annotation vectors corresponding to the $i$-th entity (i.e., $\boldsymbol h^d_{e^{start}_i},\dots,\boldsymbol h^d_{e^{end}_i}$).

During inference, we select the top $k$ entities with highest likelihood under our model. We use $k=6$ in our experiments as determined by hyper-parameter search.

\subsubsection{Pointer Networks} While a significant fraction of answers in QA datasets like SQuAD are entities, entities alone may be insufficient for detecting different aspects of a document. Many documents are entity-less, and entity taggers may fail to recognize some entities. To this end, we build a neural model that is trained from scratch to extract all human-selected answer phrases in a particular document. We parameterize this model as a pointer network \cite{vinyals2015pointer} trained to point sequentially to the start and end \emph{locations} of all labeled answers in a document. An autoregressive decoder LSTM augmented with an attention mechanism is then trained to point (attend) to all of the start and end locations of answers \emph{from left to right}, conditioned on the annotation vectors (extracted in the same fashion as in the NES model), via an attention mechanism. We add a special termination token to the document and train the decoder to attend to it once it has generated all key phrases. This enables the model to extract variable numbers of key phrases depending on the input document. This is in contrast to the work of \citet{meng2017deepkpg}, where a fixed number of key phrases is generated per document.

A pointer network is an extension of sequence-to-sequence models \cite{sutskever2014sequence}, where the target sequence consists of positions in the source sequence. An autoregressive decoder RNN is trained to attend to these positions in the input conditioned on an encoding of the input produced by an encoder RNN. We denote the decoder's annotation vectors as $(\boldsymbol h_1^p,\boldsymbol h_2^p,\dots,\boldsymbol h_{2n_{a}-1}^p,\boldsymbol h_{2n_{a}}^p)$, where $n_a$ is the number of answer key phrases, $\boldsymbol h_1^p$ and $\boldsymbol h_2^p$ correspond to the start and end annotation vectors for the first answer key phrase, and so on. We parameterize $P(w^d_i=start|\boldsymbol h_1^p \dots \boldsymbol h_j^p,\boldsymbol h^d)$ and $P(w^d_i=end|\boldsymbol h_1^p \dots \boldsymbol h_j^p,\boldsymbol h^d)$ using the general attention mechanism \cite{luong2015effective} between the decoder and encoder annotation vectors,
\begin{eqnarray*}
P(w^d_i|\boldsymbol h_1^p \dots \boldsymbol h_j^p,\boldsymbol h_{\cdot}^d) & = & \mathtt{softmax}(W_1\boldsymbol h_j^p\cdot \boldsymbol h_{\cdot}^d),
\end{eqnarray*}
where $W_1$ is a learned parameter matrix. The inputs at each step of the decoder are words from the document that correspond to the start and end locations pointed to by the decoder.

During inference, we employ a decoding strategy that greedily picks the best location from the softmax vector at every step, then post process results to remove duplicate key phrases. Since the output sequence is relatively short, we observed similar performances when using greedy decoding and beam search.

We also experimented with a BIO tagging model using an LSTM-CRF \cite{lample2016neural} but were unable to make the model predict anything except ``O'' for every token.

\subsection{Question Generation}
The question generation model adopts a sequence-to-sequence framework
\cite{sutskever2014sequence} with an attention mechanism
\cite{DBLP:journals/corr/BahdanauCB14} and a pointer-softmax decoder
\cite{gulcehre2016pointing}. We make use of the pointer-softmax mechanism since it lets us take advantage of the inherent nature of RC datasets re-using words in the document when framing questions. Our setup for this module is identical to \cite{yuan2017machine}. It takes a document $w^d_{1:n_d}$ and an answer
$w^a_{1:n_a}$ as input, and outputs a question $\hat
w^q_{1:n_q}$.

An input word $w^{\{d,a\}}_i$ is represented by concatenating its word embedding ${\boldsymbol e}_i$ with character-level embedding ${\boldsymbol e}_{ch_i}$. Each character in the alphabet receives an embedding vector, and ${\boldsymbol e}_{ch_i}$ is the final state of a bi-LSTM running over the embedding vectors corresponding to the character sequence of the word.

To leverage the extractive nature of answers in SQuAD, we encode an answer using the document annotation vectors at the answer-word positions. Specifically, if an answer phrase $w^a_{1:n}$ occupies the document span $w^d_{a_1:a_n}$, we first encode the corresponding document annotation vectors with a \emph{condition aggregation} BiLSTM into $\boldsymbol h'_{1:n}$. The concatenation of the final state $\boldsymbol h'_n$ with the answer annotation vector $\boldsymbol h^a_n$ as the answer representation.

The RNN decoder employs a pointer-softmax module
\cite{gulcehre2016pointing}. At each step of the generation process, the decoder decides adaptively whether to (a) generate from the decoder vocabulary or (b) point to a word in the source sequence (the document) and copy over.
The pointer-softmax decoder thus has two components --- a pointer attention mechanism and a generative decoder.

The subsequent mathematical notation deviates from the previous notation slightly, we use (t) as the superscript. In the \textbf{pointing decoder}, recurrence is
implemented with two cascading LSTM cells $c_1$ and $c_2$:
\begin{eqnarray}
 \label{eqn:dec_lstm1}
 {\boldsymbol s}_1^{(t)} & = & c_1({\boldsymbol y}^{(t-1)}, {\boldsymbol s}_2^{(t-1)})\\
 \label{eqn:dec_lstm2}
 {\boldsymbol s}_2^{(t)} & = & c_2({\boldsymbol v}^{(t)}, {\boldsymbol s}_1^{(t)}),
\end{eqnarray}
where ${\boldsymbol s}_1^{(t)}$ and ${\boldsymbol s}_2^{(t)}$ are the recurrent states, ${\boldsymbol y}^{(t-1)}$
is the embedding of decoder output from the previous time step,
and ${\boldsymbol v}^{(t)}$ is the context vector, which is the sum of the document annotations ${\boldsymbol h}^d_i$ weighted by the document attention $\alpha^{(t)}_i$ (Equation \eqref{eqn:alpha}):
\begin{eqnarray*}
  {\boldsymbol v}^{(t)}=\sum_{i=1}^n \alpha^{(t)}_i{\boldsymbol h}^d_i.
\end{eqnarray*}

At each time step $t$, the pointing decoder computes a distribution $\boldsymbol\alpha^{(t)}$ over the
document word positions (i.e., a document attention,
\citealt{DBLP:journals/corr/BahdanauCB14}). Each element is defined as:
\begin{eqnarray}
 \alpha^{(t)}_i = f({\boldsymbol h}^d_i, {\boldsymbol h}^a, {\boldsymbol s_1}^{(t-1)}),
  \label{eqn:alpha}
\end{eqnarray}
where $f$ is a two-layer MLP with \emph{tanh} and \emph{softmax} activation, respectively.

The \textbf{generative decoder}, on the other hand, defines a distribution over a
prescribed decoder vocabulary with a two-layer MLP $g$:
\begin{eqnarray*}
  {\boldsymbol o}^{(t)}=g({\boldsymbol y}^{(t-1)},{\boldsymbol s}_2^{(t)},{\boldsymbol v}^{(t)},{\boldsymbol h}^a).
  \label{eqn:smx_gen}
\end{eqnarray*}
Pointer-softmax is implemented by interpolating the generative and the pointing distributions:
\begin{eqnarray*}
  P(\hat w_t)\sim s^{(t)} \boldsymbol\alpha^{(t)} + (1-s^{(t)}){\boldsymbol o}^{(t)},
\end{eqnarray*}
where $s^{(t)}$ is a switch scalar computed at each time step by a
three-layer MLP $h$:
\begin{eqnarray*}
  s^{(t)}=h({\boldsymbol s}_2^{(t)},{\boldsymbol v}^{(t)},\boldsymbol{\boldsymbol\alpha}^{(t)},{\boldsymbol o}^{(t)}).
\end{eqnarray*}
The first two layers of $h$ use \emph{tanh} activation with highway connections, and the final layer uses \emph{sigmoid} activation.\footnote{We also attach the entropy of the softmax distributions to the input of the final layer, postulating
that this guides the switching mechanism by indicating the confidence of pointing vs generating. We observed an improvement in question quality with this modification.}

\begin{table}[t!]
  \caption{Model evaluation on key phrase extraction}
  \label{tab:da_table}
  \resizebox{.48\textwidth}{!}{
    \begin{tabular}{r|cccccc}
      \toprule
      & \multicolumn{3}{c}{Validation} & \multicolumn{3}{c}{Test}\\
      Models & $F1_{MS}$ & Prec. & Rec.  & $F1_{MS}$ & Prec. & Rec.\\
      \midrule
      \multicolumn{7}{c}{\textbf{SQuAD}}\\
      \midrule
\texttt{H\&S} & - & - & - & 0.292 & 0.252 & 0.403 \\
\texttt{ENT} & 0.308 & 0.249 & \textbf{0.523} & 0.347 & 0.295 & \textbf{0.547}\\
\texttt{NES} & 0.334 & 0.335 & 0.354 & 0.362 & 0.375 & 0.380\\
\texttt{PtrNet} & \textbf{0.352} & \textbf{0.387} & 0.337 & \textbf{0.404} & \textbf{0.448} & 0.387\\
      \midrule
      \multicolumn{7}{c}{\textbf{NewsQA}}\\
      \midrule
      \texttt{ENT} & 0.187 & 0.127 & \textbf{0.491} & 0.183 & 0.125 & \textbf{0.479} \\
      \texttt{PtrNet} & \textbf{0.452} & \textbf{0.480} & 0.444 & \textbf{0.435} & \textbf{0.467} & 0.427 \\
      \bottomrule
    \end{tabular}
  }
\end{table}

\section{Experiments and Results}
\subsection{Datasets}
We conduct our experiments on the SQuAD \cite{rajpurkar2016squad} and NewsQA \cite{trischler2016} datasets. These are machine
comprehension corpora consisting of over 100k crowd-sourced question-answer pairs. SQuAD contains 536 paragraphs from Wikipedia while NewsQA was created on 12,744 news articles. Simple preprocessing is performed, including lower-casing and word tokenization using \emph{NLTK}. Since the test split of SQuAD is hidden from the public, we use 5,158 question-answer pairs (self-contained in 23 Wikipedia articles) from the training set for development, and use the official development data to report test results.

\subsection{Implementation Details}
We train all models by stochastic gradient descent, with a minibatch size of 32, using the ADAM optimizer.

\subsubsection{Key Phrase Extraction} Key phrase extraction models use pretrained, 300-dimensional word embeddings generated using a word2vec extension \cite{ling2015two} and the \textit{English Gigaword 5} corpus. We used bidirectional LSTMs of 256 dimensions (128 forward and backward) to encode the document and an LSTM of 256 dimensions as our decoder in the pointer network. A dropout rate of 0.5 was used at the output of every layer in the network.

\subsubsection{Question Generation} The question decoder uses a vocabulary of the 2000 most frequent words in the training data (questions only).
This limited vocabulary is possible because the question generator may copy over out-of-vocabulary words from the document with its Pointer-Softmax mechanism.
The decoder embedding matrix is initialized with 300-dimensional \emph{GloVe} vectors \cite{pennington2014glove}, and dimensionality of the character representations is 32. The number of hidden units is 384 for both the encoder and decoder RNN cells. Dropout is applied at a rate of 0.3 to all embedding layers as well as between the hidden states in the encoder/decoder RNNs across time steps.

\begin{table*}[t!]
	\small
  \caption{Qualitative examples of detected key phrases and generated questions.}
  \label{tab:qual}
  \resizebox{.98\textwidth}{!}{
    \begin{tabular}{rp{.5\textwidth}p{.5\textwidth}}
      \toprule
        Doc. & \multicolumn{2}{p{\textwidth}}{\small\it \textbf{inflammation} is one \textbf{of the first responses of the immune system to infection} . the symptoms of inflammation are redness , swelling , heat , and pain , which are caused by increased blood flow into tissue . inflammation is produced \textbf{by eicosanoids and cytokines} , which are released by injured or infected cells . eicosanoids include prostaglandins that produce fever and the dilation of blood vessels associated with inflammation , and \textbf{leukotrienes} that attract certain white blood cells ( leukocytes ) $\dots$}\\
        % \midrule
        \multirow{4}{*}{Q-A} & \multicolumn{2}{c}{\texttt{H\&S}} \\
        & {\small\it \textbf{by eicosanoids and cytokines} --- who is inflammation produced by ?}
        & {\small\it \textbf{of the first responses of the immune system to infection} --- what is inflammation one of ?}\\
        \multirow{4}{*}{Q-A} & \multicolumn{2}{c}{\texttt{PtrNet}} \\
        & {\small\it \textbf{leukotrienes} --- what can attract certain white blood cells ?}
        & {\small\it \textbf{eicosanoids and cytokines} --- what are bacteria produced by ?}\\
        \multirow{4}{*}{Q-A} & \multicolumn{2}{c}{\texttt{Gold SQuAD}} \\
        & {\small\it \textbf{inflamation} --- what is one of the first responses the immune system has to infection ?}
        & {\small\it \textbf{eicosanoids and cytokines} --- what compounds are released by injured or infected cells , triggering inflammation ?}\\
        \midrule
        Doc. & \multicolumn{2}{p{\textwidth}}{\small\it \textbf{research} shows \textbf{that student motivation and attitudes towards school are closely linked to student-teacher relationships} . \textbf{enthusiastic teachers} are particularly good at creating \textbf{beneficial} relations with their students . their ability to create effective learning environments that foster student achievement depends on the kind of relationship they build with their students . \textbf{useful teacher-to-student interactions} are crucial in linking academic success with personal achievement . here , personal success is a student 's internal goal of improving himself , whereas academic success includes the goals he receives from his superior . 
        \textbf{a teacher} must guide his student in \textbf{aligning his personal goals with his academic goals} . students who receive this positive influence show stronger self-confidenche and greater personal and academic success than those without these teacher interactions .}\\
        % \midrule
        \multirow{4}{*}{Q-A} & \multicolumn{2}{c}{\texttt{H\&S}} \\
        & {\small\it \textbf{research} --- what shows that student motivation and attitudes towards school are closely linked to student-teacher relationships ?}
        & {\small\it \textbf{useful teacher-to-student interactions} --- what are crucial in linking academic success with personal achievement ?}\\
        & {\small\it \textbf{to student-teacher relationships} --- what does research show that student motivation and attitudes towards school are closely linked to ?}
        & {\small\it \textbf{that student motivation and attitudes towards school are closely linked to student-teacher relationships} --- what does research show to ?}\\
        \multirow{4}{*}{Q-A} & \multicolumn{2}{c}{\texttt{PtrNet}} \\
        & {\small\it \textbf{student-teacher relationships} --- what are the student motivation and attitudes towards school closely linked to ?}
        & {\small\it \textbf{enthusiastic teachers} --- who are particularly good at creating beneficial relations with their students ?}\\
        & {\small\it \textbf{teacher-to-student interactions} --- what is crucial in linking academic success with personal achievement ?}
        & {\small\it \textbf{a teacher} --- who must guide his student in aligning his personal goals ?}\\
        \multirow{4}{*}{Q-A} & \multicolumn{2}{c}{\texttt{Gold SQuAD}} \\
        & {\small\it \textbf{student-teacher relationships} --- 'what is student motivation about school linked to ?}
        & {\small\it \textbf{beneficial} --- what type of relationships do enthusiastic teachers cause ?}\\
        & {\small\it \textbf{aligning his personal goals with his academic goals .} --- what should a teacher guide a student in ?}
        & {\small\it \textbf{student motivation and attitudes towards school} --- what is strongly linked to good student-teacher relationships ?}\\
        \midrule
        Doc. & \multicolumn{2}{p{\textwidth}}{\small\it \textbf{the yuan dynasty} was \textbf{the first time} that \textbf{non-native chinese people} ruled all of china . in the historiography of mongolia , it is generally considered to be the continuation of \textbf{the mongol empire} . mongols are widely known to \textbf{worship the eternal heaven} $\dots$} \\
        \multirow{4}{*}{Q-A} & \multicolumn{2}{c}{\texttt{H\&S}} \\
        & {\small\it \textbf{the first time} --- what was the yuan dynasty that non-native chinese people ruled all of china ?}
        & {\small\it \textbf{the yuan dynasty} --- what was the first time that non-native chinese people ruled all of china ?}\\
        \multirow{4}{*}{Q-A} & \multicolumn{2}{c}{\texttt{PtrNet}} \\
        & {\small\it \textbf{the mongol empire} --- the yuan dynasty is considered to be the continuation of what ?}
        & {\small\it \textbf{worship the eternal heaven} --- what are mongols widely known to do in historiography of mongolia ?}\\
        \multirow{4}{*}{Q-A} & \multicolumn{2}{c}{\texttt{Gold SQuAD}} \\
        & {\small\it \textbf{non-native chinese people} --- the yuan was the first time all of china was ruled by whom ?}
        & {\small\it \textbf{the eternal heaven} --- what did mongols worship ?}\\
        \midrule
        Doc. & \multicolumn{2}{p{\textwidth}}{\small\it on july 31 , 1995 , the walt disney company announced an agreement to merge with capital cities/abc for \$ 19 billion . $\dots$ . in 1998 , abc premiered the aaron sorkin-created sitcom sports night , centering on the travails of the staff of a sportscenter-style sports news program ; despite earning critical praise and multiple emmy awards , the series was cancelled in 2000 after two seasons .} \\
        \multirow{4}{*}{Q-A} & \multicolumn{2}{c}{\texttt{H\&S}} \\
        & {\small\it \textbf{an agreement to merge with capital cities/abc for \$19 billion} --- what did the walt disney company announce on july 31 , 1995 ?}
        & {\small\it \textbf{the walt disney company} --- what announced an agreement to merge with capital cities/abc for \$19 billion on july 31 , 1995 ?}\\
        \multirow{4}{*}{Q-A} & \multicolumn{2}{c}{\texttt{PtrNet}} \\
        & {\small\it \textbf{2000} --- in what year was the aaron sorkin-created sitcom sports night cancelled ?}
        & {\small\it \textbf{walt disney company} --- who announced an agreement to merge with capital cities/abc for \$ 19 billion ?}\\
        \multirow{4}{*}{Q-A} & \multicolumn{2}{c}{\texttt{Gold SQuAD}} \\
        & {\small\it \textbf{july 31 , 1995} --- when was the disney and abc merger first announced ?}
        & {\small\it \textbf{sports night} --- what aaron sorkin created show did abc debut in 1998 ?}\\
      \bottomrule
    \end{tabular}
  }
\end{table*}

\subsection{Quantitative Evaluation of Key Phrase Extraction}
Since each key phrase is itself a multi-word unit, we believe that a na\"ive, word-level \emph{F1} that considers an entire key phrase as a single unit is not well suited to this evaluation. We therefore propose an extension of the SQuAD F1 metric (for a single answer span) to multiple spans within a document, which we call the \emph{multi-span F1 score}.

This metric is calculated as follows. Given the predicted phrase $\hat e_i$ and a gold phrase $e_j$, we first construct a pairwise, token-level $F1$ score matrix of elements $f_{i,j}$ between the two phrases $\hat e_i$ and $e_j$. Max-pooling along the gold-label axis essentially assesses the precision of each prediction, with partial matches accounted for by the pair-wise \emph{F1} (identical to evaluation of a single answer in SQuAD) in the cells: $p_{i}=\max_j(f_{i,j})$. Analogously, the recall for label $e_j$ can be computed by max-pooling along the prediction axis: $r_{j}=\max_i(f_{i,j})$. We define the \emph{multi-span F1 score} using the mean precision $\bar p=\mathrm{avg}(p_{i})$ and recall  $\bar r=\mathrm{avg}(r_{j})$:
\[F1_{MS}=\frac{2 \bar p\cdot\bar r}{(\bar p+\bar r)}.\]

Note that existing evaluations (e.g., that of \citet{meng2017deepkpg}) can be seen as the above computation performed on the matrix of exact match scores between predicted and gold key phrases. By using token-level F1 scores between phrase pairs, we allow fuzzy matches.

Our evaluation of key phrase extraction systems by this metric is presented in Table~\ref{tab:da_table}. We compare answer phrases extracted by the method of \citet{heilman2010good} (henceforth refered to as H\&S),\footnote{http://www.cs.cmu.edu/~ark/mheilman/questions/} our baseline entity tagger, the neural entity selection module, and the pointer network. As expected, the entity tagging baseline achieves the best recall, likely by over-generating candidate answers. The \texttt{NES} model, on the other hand, exhibits much better precision and consequently outperforms the entity tagging baseline significantly in \emph{F1}. This trend persists when comparing the \texttt{NES} model and the pointer network. The H\&S model exhibits high recall but lacks precision, similar to the baseline entity tagger. This is not surprising since that model is not trained on SQuAD's answer-phrase distribution.

\subsection{Qualitative Evaluation of Key Phrase Extraction}
Qualitatively, we observe that the entity-based models have a strong bias for numeric types, which often fail to capture interesting information in a document. We also notice that entity-based systems tend to select the central topical entity as answer, which does not match the distribution of answers typically selected by humans. For example, given a Wikipedia article on Kenya stating that \emph{agriculture is the second largest contributor to kenya 's gross domestic product (gdp)}, entity-based systems propose \emph{kenya} as an answer phrase. This leads to the (low-quality) question \emph{what country is nigeria's second largest contributor to?} \footnote{Since the answer word \emph{kenya} cannot appear in the generated question, the decoder produced a similar word \emph{nigeria} instead.} Given the same document, the pointer model picked \emph{agriculture} as the answer and asked \emph{what is the second largest contributor to kenya 's gross domestic product ?}

\subsection{Quantitative Evaluation of QA pairs}
We can quantitatively evaluate our question generation module by conditioning it on gold answers from the SQuAD development set. We can then use standard automatic evaluation metrics for generative models of text such as BLEU.
Our question generation model evaluated in such a manner yields 10.4 BLEU$_4$.

However, there can exist a many possible ways to formulate a question given the same answer. BLEU thus becomes a less desirable metric by penalizing any generation that does not closely match (lexically) the reference question. To address this issue, we propose to evaluate a generated question by employing a pre-trained QA model. Specifically, suppose question $\hat q$ is generated from document $d$ and answer $a$, and the pre-trained QA model outputs answer $\hat a$ given the input $d$ and $\hat q$. If the QA model is assumed to be able to answer the gold question $q$ with the gold answer $a$, then the F1 score between $a$ and $\hat a$ may serve as a proxy to the \textit{semantic} equivalence between $q$ and $\hat q$ --- regardless of the amount of word/n-gram overlap between $q$ and $\hat q$.

Quantitatively, a match-LSTM model \cite{wang2016machine} pre-trained on gold squad question/answer pairs achieves an F1 score of 72.4\% on our generated questions in comparison to 73.8\% on the SQuAD dev set.

In addition to the automatic evaluation metrics, we also undertook a human evaluation of generated questions and answers.

\subsection{Qualitative Evaluation of QA pairs}
We present several answer-extraction and question-generation examples in Table~\ref{tab:qual}. Each example contains a document and three corresponding QA pairs, generated respectively by H\&S,
by our two-stage framework, and by the original SQuAD crowdworkers.

We now discuss the relative qualities of QA pairs from each synthetic method.

\begin{figure*}[ht!]
\centering
\includegraphics[scale=0.5]{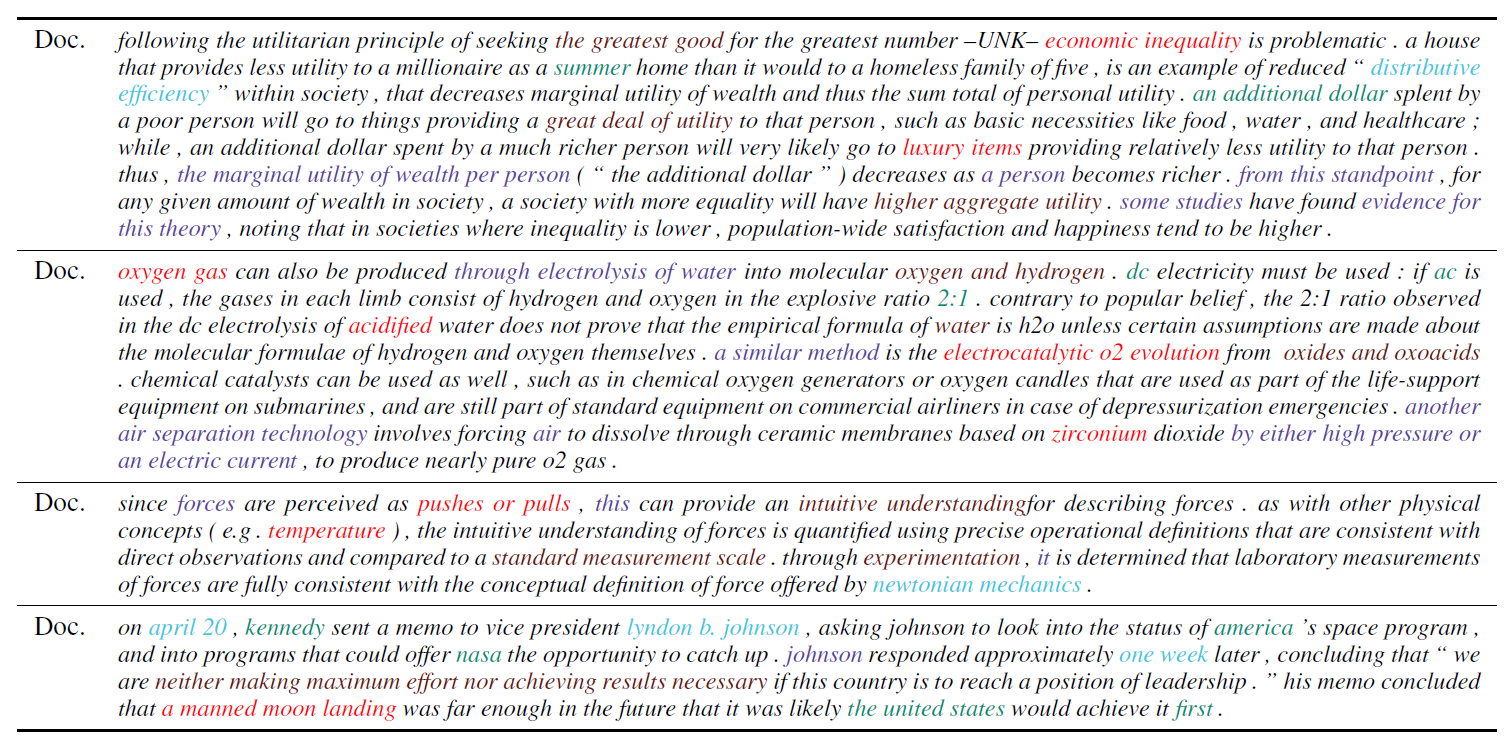}
\caption{A comparison of key phrase extraction methods. Red phrases are extracted by the pointer network, violet by H\&S, green by the baseline, brown correspond to squad gold answers and cyan indicates an overlap between the pointer model and squad gold questions. The last paragraph is an exception where \textbf{lyndon b. johnson} and \textbf{april 20} are extracted by H\&S as well as the baseline model.}
\label{fig:keyphrasecompare}
\end{figure*}

\paragraph{H\&S} Key phrases selected by the H\&S model are structurally distinct from the PtrNet and human-generated answers. For example, they may start with prepositions, such as \emph{of}, \emph{by}, and \emph{to}, or consist of very long phrases like \emph{that  student  motivation  and  attitudes  towards  school  are  closely linked to student-teacher relationships}. As seen in Figure \ref{fig:keyphrasecompare}, these key phrases may also contain vague phrases such as ``this theory'', ``some studies'', ``a person'', etc., which renders them less natural for question generation. The H\&S question generator appears to produce a few ungrammatical sentences, e.g., \textit{the first time -- what was the yuan dynasty that non-native chinese people ruled all of china ?}

\paragraph{Our system} Since our key phrase extractor was trained on SQuAD, the selected key phrases more closely resemble gold SQuAD answers. However, sometimes the generated questions do not target the extracted answers, eg, \textit{eicosanoids and cytokines --- what are bacteria produced by ?} (first document in Table~\ref{tab:qual}). Interestingly, our model is sometimes able to resolve coreferent entities. For instance, to generate \textit{the mongol empire -— the yuan dynasty is considered to be the continuation of what ?} the model must resolve the pronoun \textit{it} to \textit{yuan dynasty} in \textit{it is generally considered to be the continuation of the mongol empire} (third document in Table~\ref{tab:qual}).

\subsection{Human Evaluation Studies}
We carried out human evaluations on the question generation module in isolation as well as in conjunction with the key phrase extraction module. 

\paragraph{Evaluating the ability of the Question Generation Module to transfer to new settings}
We asked crowdworkers part of an internal evaluation system to evaluate two different aspects of questions generated by our module - fluency and correctness. Our system was provided Internet articles and candidate answers selected from an internal search engine thereby evaluating the model's ability to generalize from simple RC datasets to the real world. For fluency evaluations, they were asked whether the generated questions sounded natural (ignoring semantics) with scores of 0/1/2 corresponding to "No", "Somewhat" and "Yes". 17.5\% were labeled 0, 22.7\% were labeled 1 and 59.8\% were labeled 2. For correctness evaluations, annotators were asked if the given answer was the correct answer for the given question. 64.4\% of questions were labeled incorrect, leaving 35.6\% labeled as correct. This particular evaluation differs slightly from others with regard to the module used (it was trained a combination of SQuAD + NewsQA + TriviaQA \cite{joshi2017triviaqa}). Also the documents and answers used provided via an internal tool. 1,302 annotations were collected.

\paragraph{Comparison to human generated questions} We present annotators with documents from SQuAD's official development set and two sets of question-answer pairs, one from our model (machine generated) and the other from SQuAD (human generated). Annotators are then asked to identify which question-answer pair is machine generated. The order in which the pairs appear is randomized across examples. Annotators are free to use any criterion to make a distinction, such as poor grammar, the answer phrase not correctly answering the generated question, unnatural answer phrases, etc.

We presented 14 annotators with a total of 740 documents, each with 2 corresponding QA pairs. Annotators identified the machine generated pairs 77.8\% of the time with a standard deviation of 8.34\%.

\paragraph{Implict comparison to H\&S} To compare our system to existing methods (H\&S), we orchestrate an implict comparison grounded in human generated QA pairs from SQuAD. We present human annotators with a document and two QA pairs -- one that comes from the true development set and the other from either our system or H\&S, at random. Annotators are not told that there are two different models generating QA pairs. As above, annotators are asked to identify which QA pair is human generated and which is synthetic.

We presented a single annotator with 100 documents, each with two QA pairs. For 45 documents, the synthetic QA pair came from from our model; for the remaining 55, the synthetic pair was from H\&S. The annotator distinguished correctly between our system's output and the human-generated pair in 30 cases (66.7\%), and did so in 45 cases (81.8\%) for H\&S. This experiment suggests that our system's generated QA pairs are less distinguishable from human QA pairs.

\paragraph{Comparison to H\&S} In a more direct evaluation, we present annotators with documents from the SQuAD development set along with one QA pair generated by the H\&S model and one generated by ours. We then ask annotators which QA pair they prefer.

We presented the same single annotator with 200 such examples. In 107 cases (53.5\%), the annotator preferred the pair generated by our model. This suggests that, without human generated QA pairs for comparison, the annotator considers the two models' outputs to be roughly equal in quality.

\section{Conclusion}
We propose a two-stage framework to tackle the problem of question generation from documents. First, we use a question answering corpus to train a neural model to estimate the distribution of key phrases that are likely to be picked by humans to ask questions about. We present two neural models, one that ranks entities proposed by an entity tagging system, and another that points to key-phrase start and end boundaries with a pointer network. When compared to an entity tagging baseline, the proposed models exhibit significantly better results.

We adopt a sequence-to-sequence model to generate questions conditioned on the key phrases selected in the framework's first stage. Our question generator is inspired by an attention-based translation model, and uses the pointer-softmax mechanism to dynamically switch between copying a word from the document and generating a word from a vocabulary. Qualitative examples show that the generated questions exhibit both syntactic fluency and semantic relevance to the conditioning documents and answers, and appear useful for assessing reading comprehension in educational settings. In future work we will investigate fine-tuning the two-stage framework end to end. Another interesting direction is to explore abstractive key-phrase extraction.

\section{Acknowledgements}
The authors would like to thank the reviewers for their valuable feedback.

\bibliography{acl2018}
\bibliographystyle{acl_natbib}

\end{document}